# Robotic Path Planning Implementation using Search Algorithms

First A. Vikram Shahapur, Second B. Blessing Dixon, Third C. Urvishkumar Bharti

*Abstract*— Till now, many path planning algorithms have been proposed in the literature. The objective of these algorithms is to find the quickest path between initial position to the end position in a certain environment. The complexity of these algorithms depends on the internal parameters such as motor speed or sensor range and on other external parameters, including the accuracy of the map, size of the environment, and the number of obstacles. In this paper, we are giving information about how path planning algorithm finds the optimal path in an uneven terrain with a multiple obstacle using TurtleBot3 robot into the Gazebo environment using Dijkstra's and A*.

## I. INTRODUCTION

A fundamental task for any mobile robot is its capability to organize collision-free trajectories from point A to point B, a start to a end position or visiting a series of positions, i.e. regions of interest. When provided a map and a end point, path planning includes selecting the best (collision free, if applicable) trajectory that the robot can follow to reach the goal position. Ultimately, this is a problem of finding the optimal subset from a set of possible trajectories that robot could follow while transitioning to the target location.

For the robot to be Reliable and effective in an environment, an efficient path planning algorithm is needed. In this project we discuss and implement the A* (pronounced "A-star") search-based algorithm using ROS. Search algorithms are widely used to solve problems that can be modeled as a graph. The quality of the produced path affects immensely the robotic application, because in the worst cases scenario most of the path planning algorithms won't show the optimal path. Usually, the minimization of the covered distance is the primary aim of the navigation process as it impacts the other system of measurement such as the dealing with time and the power consumption.

Planning is obvious part of navigation that answers the question: What is the best way to go there? A robot must be able to get to the target point while avoiding the obstacles in the environment with optimal path length. There are numerous concerns need to be considered in the path planning of mobile robots due to a variety of objectives and tasks of the practical robot itself. Nearly all the suggested methodologies are concentrating on locating the shortest path from the primary spot to end point. Figure 1 demonstrates a conventional navigation architecture of a mobile robot. As observed in Figure 1, there are four sections in the architecture [1]. Among all the layers, path planning is the first layer, as a result accuracy of path planning affects the application of the robot. The second layer in this navigation architecture comprises the motion controller which is responsible for generating the control actions in such a means that the robot follows as precisely as possible the trajectory generated by the path planning algorithms.

The third layer is the low-level PID controllers that make certain that the control actions generated by the motion controller are reached by the robot actuators and sensors. It is essential for the motion controller to know the current location of the robot. This is determined by the localization layer in which multiple types of sensors are used to get the robot's position.

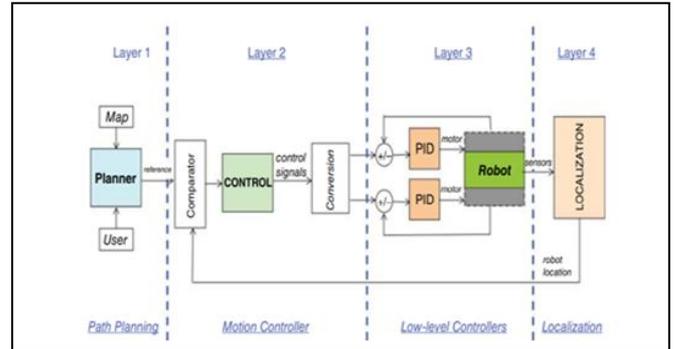

Figure 1. Mobile Robot Navigation Architecture

## I. PROBLEM FORMULATION

The focus of this project is path-planning, which involves finding the optimal or near-optimal path from point A to point B. The aim here is to find the shortest path, optimally, that minimizes computation and maximizing the efficiency of the robot. That is, we would like the robot to get to a goal state while using the least amount of energy possible.

Planning a pathway in large-scale environments is more difficult as the trouble turn out to be further complex and time-consuming which is not suitable for robotic applications in which real-time aspect is crucial [1].

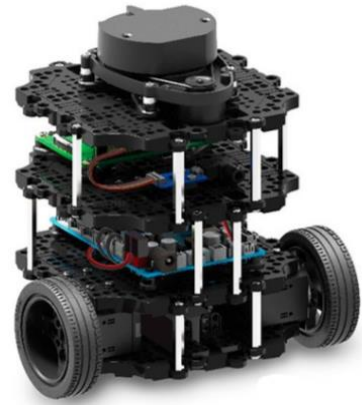

Figure 2. Differential Drive Turtlebot

## II. PROPOSED SOLUTION

For this problem we would first like to understand the given environment, understand the robot dynamics, and apply both the environment model and the robot dynamic to a well-suited algorithm, A* search-based algorithm in this case. Figure 2 shows the TurtleBot3 which we have used in our simulation for path planning. TurtleBot3 is one of the types of differential drive mobile robot. Here we are implementing TurtleBot3 into the known environment with multiple obstacles. After generating the map of the environment, we have implemented the path planning algorithm to find out the shortest path between the current location to the end location. Remember that, for path planning, we must know the current and goal

position of the bot. Below are details on the different components involve in solving this problem:

### A. The Environment Map

We first would like to know the map of the environment. A map of the environment is essential to solving this problem. The map provides a foundation or guide or a frame for referencing the robot pose. Using this environment map, we can generate a state/node-based graph representation of the environment. This graph provides us with a low-level deterministic understanding of all possible occupancy in the environment. However, for the graph to be consistent or representative of possible actions of the robot, it must be generated using the state-action simulation of the robot. This is where the robot's *state-space model* comes in handy.

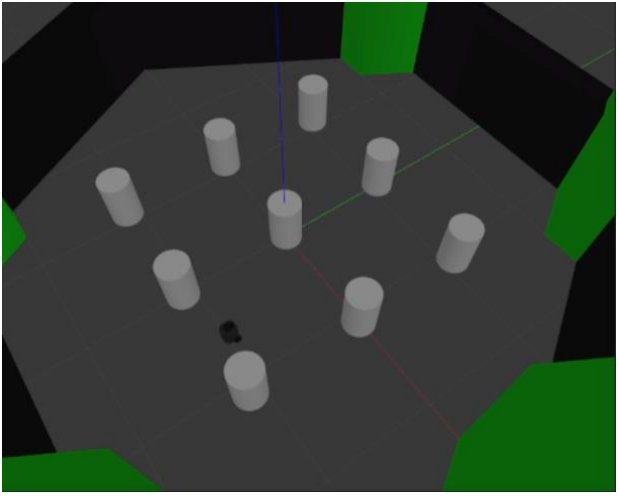

Figure 3. The Gazebo Environment

### B. Robot State Estimation Model

For the python 2D simulation, the accuracy of the state-space model plays a very important role in the accuracy of predicting the robot's pose within the environment at simulated time step. That is, we do not want to predict a state of the robot that varies drastically from the observable state within the environment. The use of an observation and the Extended Kalman Filter can greatly improve the confidence of estimation. Below are the equations describing the state a differential drive robot model:

i. Non-linear Model

$$\begin{bmatrix} x_t \\ y_t \\ \gamma_t \end{bmatrix} = \begin{bmatrix} x_{t-1} + v_{t-1} cos\gamma_{t-1} * dt \\ y_{t-1} + v_{t-1} sin\gamma_{t-1} * dt \\ \gamma_{t-1} + \omega_{t-1} * dt \end{bmatrix} = \begin{bmatrix} f_1 \\ f_2 \\ f_3 \end{bmatrix}$$

$$\begin{bmatrix} x_t \\ y_t \\ \gamma_t \end{bmatrix} = A_{t-1} \begin{bmatrix} x_{t-1} \\ y_{t-1} \\ \gamma_{t-1} \end{bmatrix} + B_{t-1} \begin{bmatrix} v_{t-1} \\ \omega_{t-1} \end{bmatrix}$$

$$\boldsymbol{x_t} = A_{t-1}\boldsymbol{x_{t-1}} + B_{t-1}\boldsymbol{u_{t-1}}$$

$x_t -$ *obot's position on the x - axis*
$y_t -$ *robot's position on the y - axis*
$\gamma_t -$ *robot's rotation in radians*

$v_{t-1} -$ *linear velocity of the robot*
$\omega_{t-1} -$ *angular velocity of the robot*

ii. Linearized Model

$$A_{t-1} = \begin{bmatrix} \frac{\partial f_1}{\partial x_{t-1}} & \frac{\partial f_1}{\partial y_{t-1}} & \frac{\partial f_1}{\partial \gamma_{t-1}} \\ \frac{\partial f_2}{\partial x_{t-1}} & \frac{\partial f_2}{\partial y_{t-1}} & \frac{\partial f_2}{\partial \gamma_{t-1}} \\ \frac{\partial f_3}{\partial x_{t-1}} & \frac{\partial f_3}{\partial y_{t-1}} & \frac{\partial f_3}{\partial \gamma_{t-1}} \end{bmatrix}$$

iii. Observation Model

$$\boldsymbol{O_t} = H_t \boldsymbol{x_t} + \boldsymbol{w_t}$$

$$\begin{bmatrix} O_1 \\ O_2 \\ O_3 \end{bmatrix} = H_t \begin{bmatrix} x_t \\ y_t \\ \gamma_t \end{bmatrix} + \begin{bmatrix} w_1 \\ w_2 \\ w_3 \end{bmatrix}$$

$\boldsymbol{O_t} -$ *observed state*
$\boldsymbol{w_t} -$ *added sensor noise*

The assumption here is that the sensor can give the exact robot position at each time step.

Using trigonometry, we can get the following equations for a range r and bearing b:

$$r = \sqrt{(x_t - x_{landmark})^2 + (y_t - y_{landmark})^2}$$

$$b = atan2(y_t - y_{landmark}, x_t - x_{landmark}) - \gamma_t$$

$$\begin{bmatrix} r \\ b \end{bmatrix} = \begin{bmatrix} h_1 \\ h_2 \end{bmatrix}$$

Linearization of the sensing model:

$$H_t = \begin{bmatrix} \frac{\partial r}{\partial x_{t-1}} & \frac{\partial r}{\partial y_{t-1}} & \frac{\partial r}{\partial \gamma_{t-1}} \\ \frac{\partial b}{\partial x_{t-1}} & \frac{\partial b}{\partial y_{t-1}} & \frac{\partial b}{\partial \gamma_{t-1}} \end{bmatrix}$$

$$H_t = \begin{bmatrix} \frac{x_t - x_{landmark}}{\sqrt{(x_t - x_{landmark})^2 + (y_t - y_{landmark})^2}} & \frac{y_t - y_{landmark}}{\sqrt{(x_t - x_{landmark})^2 + (y_t - y_{landmark})^2}} & 0 \\ \frac{-(y_t - y_{landmark})}{(x_t - x_{landmark})^2 + (y_t - y_{landmark})^2} & \frac{x_t - x_{landmark}}{(x_t - x_{landmark})^2 + (y_t - y_{landmark})^2} & -1 \end{bmatrix}$$

### C. Generated Graph

After the graph representation of the environment is generated, the node within the graph that is closest to the target's location is designated as the *goal* node. Using this goal node, the initial starting point of the robot, and the graph itself, an appropriate search-based algorithm is implemented.

*D. The Search Algorithm*

Search algorithms are widely used in solving search problems. Search problems represents a binary relationship between entities [2]. That is, the goal of the problem is to find a structure 's' in an object 'o', and an algorithm solves this problem by finding at least one corresponding structure [3]. Such problems occur commonly in graph theory, for example, searching a graph for matches, cliques, or an independent set [3]. For a problem to be consider a search problem, the following must be defined:

1. Set of states -a structure of nodes/locations
2. Start state -an initial node/location
3. Goal state -a target node/location
4. A successor function -a mapping to neighbors

The search algorithm of choice in this project was the A* algorithm, shown in figure 4 [4,6]. The algorithm is a variant of the best-first algorithm. The goal of the A* algorithm is to find a path to a target node that has the minimum cost. In this sense, a graph can be thought of as a tree of paths from a root (or staring) node extending one edge at a time.

The A* algorithm utilizes a heuristic method to obtain optimality and completeness. The property of optimality and completeness guarantees that the algorithm will find the best possible solution for the problem, if one exists. In the case of this project, the best possible solution is the maximumly costing path to the target.

*E. Challenges*

- One of the challenges in this project was the Formulation of the problem into the search-based domain. That is, coming up with a best way to model the problem so that it can be modeled as a graph.
- There were a few challenges in writing the python algorithm for recursively simulating the robot states/nodes in the environment for graph-map generation.
- There were also challenges in plotting and simulating in the python 2D matplotlib environment
- We had challenges in setting up the ROS gazebo environment for simulating the A* algorithm using Turtlebot.

III. RESULTS

Results for simulation and experiment showed the efficiency and consistency of the A* algorithm. The python 2D environment was first used for implementation and testing of this project. An environment of width=100x and length=100 was used to simulate the robot state and generated a 2D environment graph representation of the robot state. We later extended this implementation to the ROS Gazebo environment using Turtlebot3.

For the ROS implementation using Turtlebot3, we first generated a map of the environment using SLAM navigation, for which we used the g-mapping package available in ROS. Once we have the generated map, we loaded the map onto the visualization tool known as RVIZ. We assigned the end location into the map. The robot then implemented the algorithm which started to calculate the shortest path between the current location to the end location. Here we have compared the Dijkstra's and A* algorithm.

```
// A* Search Algorithm
1.  Initialize the open list
2.  Initialize the closed list
    put the starting node on the open
    list (you can leave its f at zero)

3.  while the open list is not empty
    a) find the node with the least f on
       the open list, call it "q"

    b) pop q off the open list

    c) generate q's 8 successors and set their
       parents to q

    d) for each successor
        i) if successor is the goal, stop search

        ii) else, compute both g and h for successor
            successor.g = q.g + distance between
                          successor and q
            successor.h = distance from goal to
            successor (This can be done using many
            ways, we will discuss three heuristics-
            Manhattan, Diagonal and Euclidean
            Heuristics)

            successor.f = successor.g + successor.h

        iii) if a node with the same position as
             successor is in the OPEN list which has a
             lower f than successor, skip this successor

        iV) if a node with the same position as
            successor is in the CLOSED list which has
            a lower f than successor, skip this successor
            otherwise, add the node to the open list
        end (for loop)

    e) push q on the closed list
    end (while loop)
```

Figure 4. Pseudo code of the A* algorithm

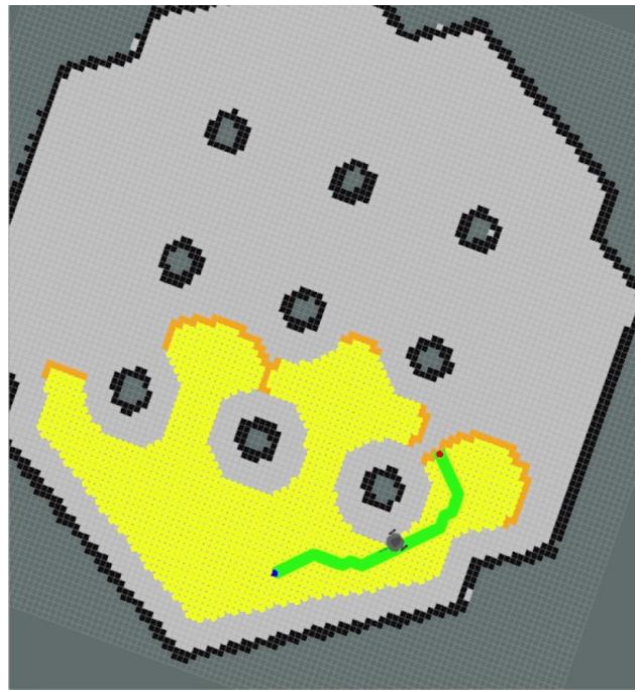

Figure 5. Implementation of Dijkstra's algorithm

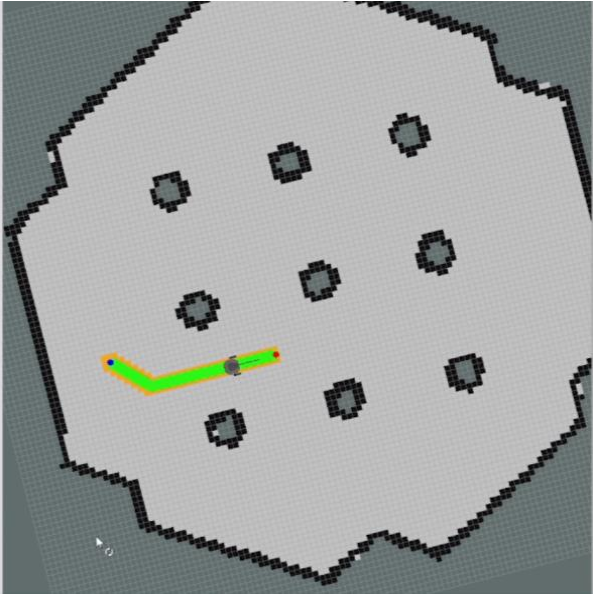

Figure 6. Implementation of A* algorithm

Figure 5 shows the simulation result of the Dijkstra algorithm. We can observe from the figure that, Dijkstra found the shortest path and the Turtlebot is following that path. Figure 6 shows the simulation results of the A* algorithm. Since A* have the heuristics function, it is finding the shortest path faster than the Dijkstra's. We can see that, A* is covering less map to find the optimal path, as a result it is more efficient than the Dijkstra's. Dijkstra's do not include the heuristic function whereas A* do. A* is combining the shortest path between the current node to the source node and approximation distance between the current node to the goal node. As a result, there is a less computation and more efficiency in A* as compared to the Dijkstra's.

## IV. CONCLUSION

In this paper, Simultaneous Localization and Mapping and Path Planning interfaced with Turtle-Bot3 to allow for safe and fast navigation from the initial position to end position. It introduced a Mapping and Planning pipeline for ground robot navigation in generated map space. A* introduces an optimal path generation with heuristic function for increasing the efficiency of the mobile robot. When compared to Dijkstra's algorithm, A* shows the high success rates which find its path in a short amount of time. As a result, this allows the robot to consume the least amount of energy. Moreover, TurtleBot3 is avoiding safely the obstacles that exist in the environment and following its optimal and obstacle-free route accurately.